\documentclass[times, review, 10pt]{elsarticle}
\usepackage{amssymb}
\usepackage{amsmath}
\usepackage[linesnumbered,ruled]{algorithm2e}
\usepackage{enumitem}
\usepackage{multirow}
\usepackage{array}
\usepackage{booktabs}
\usepackage{wrapfig}
\usepackage{graphicx}
\usepackage[caption=false,font=normalsize,labelfont=sf,textfont=sf]{subfig}
\usepackage[utf8]{inputenc}
\usepackage{setspace}
\usepackage{tikz}
\usetikzlibrary{arrows.meta,positioning,calc}

\journal{}

\begin{document}

\begin{frontmatter}

\title{Generative Verification: An Independent Signal for Active Learning of Object Detection}

\author[1]{Licheng Zhang\corref{cor1}}
\ead{licheng.zhang@student.unimelb.edu.au}
\cortext[cor1]{Corresponding author.}

\affiliation[1]{organization={School of Computing and Information Systems, The University of Melbourne},
            city={Melbourne},
            postcode={VIC 3010},
            country={Australia}}

\author[2]{Zheng Gong}
\ead{zheng.gong@jmu.edu.cn}

\affiliation[2]{organization={School of Computer Engineering, Jimei University},
            city={Xiamen},
            postcode={361021},
            state={Fujian},
            country={China}}

\begin{abstract}
Nearly every acquisition function for active object detection shares one arrangement, in that the model being improved is also the model being interrogated. We depart from it. In generative verification an independent generative model re-derives the label of a detection from the pixels inside its predicted box, and the disagreement between the two becomes the acquisition signal. Two properties follow from the arrangement itself rather than from any tuning. A displaced box, a box on background and a correct box carrying the wrong label all yield a crop that fails verification, so the failure modes arrive already combined in one scalar and the hand-weighted classification and localization terms of existing criteria are no longer needed. And because the verifier never observes the detector confidence, confidently wrong detections score highest, although a self-derived signal reads them as uninteresting and they are the costliest to leave unlabeled. We build the verifier as a conditional diffusion model whose diffusion target is a label representation rather than an image. Its reverse process is stochastic, so repeated generations return a distribution whose concentration reports how firmly the evidence determines the label, where a classifier returns a single point estimate. On PASCAL VOC and MS-COCO the signal outperforms output-uncertainty, feature-geometry, perturbation and ensemble criteria, gaining about one mAP50 point per round on MS-COCO, with its largest margins in the early rounds where confident detector errors are most common.
\end{abstract}


\end{frontmatter}

\section{Introduction}
\label{sec1}
Object detection underpins a wide range of deployed vision systems, and progress in the task has been driven by deep learning \cite{ref303}. The practical constraint on a new deployment is rarely the architecture. It is the cost of annotating data from the setting the detector must operate in \cite{ref304}. Annotating a detection dataset is far more expensive than annotating a classification dataset, since a single image requires a set of tightly drawn boxes rather than one categorical label, and the cost recurs with every new domain. Active learning addresses the cost directly by choosing which images are worth annotating, and it has been applied to image classification \cite{ref305,ref306}, video analysis \cite{ref307,ref308} and object detection \cite{ref9,zhang}.

What distinguishes one active learning method from another is almost entirely the definition of the score assigned to an unlabeled image, usually called the acquisition or informativeness function. The literature on detection has produced a long line of increasingly refined acquisition functions. Our starting point is an observation not about any one of them but about what they share. In nearly all existing work the model being improved is also the model being interrogated. Entropy \cite{ref03}, loss prediction \cite{ref5}, Gaussian mixture outputs \cite{ref9}, evidential heads \cite{ref12}, multiple instance formulations \cite{ref10}, ensembles \cite{ref6} and Monte Carlo dropout \cite{ref18} all read their signal, directly or through an auxiliary head, out of the detector itself. Two difficulties follow from the arrangement rather than from any particular acquisition function.

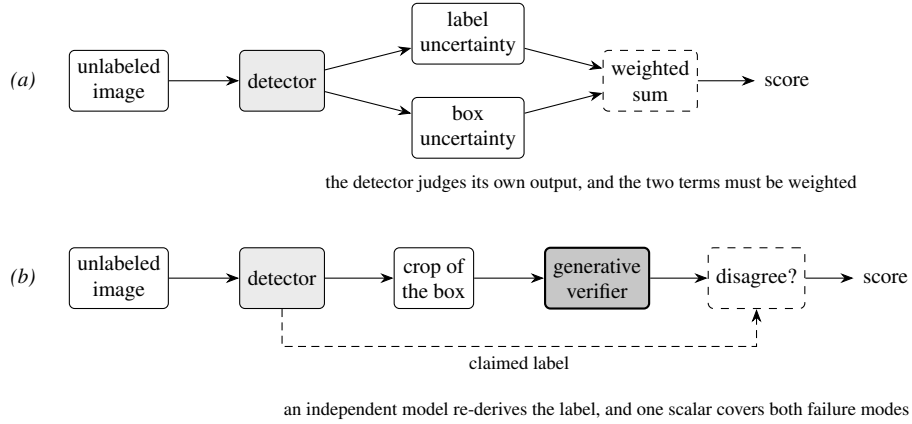
\begin{figure}[t]
\centering
\resizebox{\linewidth}{!}{%
\begin{tikzpicture}[
  font=\footnotesize,
  bx/.style={draw, rounded corners=2pt, align=center, inner sep=3pt, minimum height=8mm},
  det/.style={bx, fill=black!8},
  ver/.style={bx, fill=black!22, thick},
  fus/.style={bx, dashed},
  >={Stealth[length=5pt]},
  x=1cm, y=1cm
]
\node[bx]  (ai) at (0,0)      {unlabeled\\image};
\node[det] (ad) at (2.15,0)   {detector};
\node[bx]  (ac) at (4.6,0.62) {label\\uncertainty};
\node[bx]  (ab) at (4.6,-0.62){box\\uncertainty};
\node[fus] (af) at (7.0,0)    {weighted\\sum};
\node      (ao) at (8.8,0)    {score};
\draw[->] (ai)--(ad);
\draw[->] (ad)--(ac);
\draw[->] (ad)--(ab);
\draw[->] (ac)--(af);
\draw[->] (ab)--(af);
\draw[->] (af)--(ao);
\node[anchor=west, font=\footnotesize\itshape] at (-1.55,0) {(a)};
\node[anchor=west, font=\scriptsize] at (2.6,-1.35)
      {the detector judges its own output, and the two terms must be weighted};

\node[bx]  (bi) at (0,-2.6)    {unlabeled\\image};
\node[det] (bd) at (2.15,-2.6) {detector};
\node[bx]  (bc) at (4.15,-2.6) {crop of\\the box};
\node[ver] (bv) at (6.3,-2.6)  {generative\\verifier};
\node[fus] (bg) at (8.4,-2.6)  {disagree?};
\node      (bo) at (10.1,-2.6) {score};
\draw[->] (bi)--(bd);
\draw[->] (bd)--(bc);
\draw[->] (bc)--(bv);
\draw[->] (bv)--(bg);
\draw[->] (bg)--(bo);
\draw[->,densely dashed] (bd.south) -- (2.15,-3.5) -- node[below,font=\scriptsize] {claimed label} (8.4,-3.5) -- (bg.south);
\node[anchor=west, font=\footnotesize\itshape] at (-1.55,-2.6) {(b)};
\node[anchor=west, font=\scriptsize] at (2.05,-4.35)
      {an independent model re-derives the label, and one scalar covers both failure modes};
\end{tikzpicture}}
\caption{Two arrangements, and what separates them. (a) The prevailing one interrogates the detector it is improving, so a label term and a box term have to be combined by hand-tuned weights, and a detection the detector is confidently wrong about yields no signal. (b) Generative verification re-derives the label from the pixels inside the predicted box using a model that never sees the detector confidence. Every way a detection can fail then yields a crop that fails verification, so the failure modes arrive already combined and confident errors score highest.}
\label{fig:teaser}
\end{figure}

The first is that the informativeness of a detection has to be assembled from parts. A detection is wrong if its label is wrong or if its box is wrong, and the two predictions come from different output branches, so a score for the detection must combine a classification term with a localization term. Kao et al. \cite{ref4} combine the two uncertainties explicitly, Choi et al. \cite{ref9} fit output models to both branches, and Yu et al. \cite{ref11} require consistency of the box and the class distribution jointly. That the combination is genuinely awkward is visible in recent work moving in the opposite direction, with Zhang et al. \cite{delr} decoupling the localization query from the recognition query precisely because a detection can be informative for one and uninformative for the other. Whichever direction is taken, the relative weight of the two terms has to be tuned, and no fixed weighting is obviously correct across datasets or across rounds.

The second difficulty is more fundamental. A detector that is wrong and confident about it produces exactly the output pattern that a self-derived signal reads as uninteresting. The case is not rare and it is the most expensive one, since a confident false positive costs precision directly. Perturbation-based methods such as \cite{zhang} partially address it by testing whether the detector output is stable under a change to its own features, but the detector remains the judge of its own work.

We change the arrangement rather than the acquisition function, as Figure \ref{fig:teaser} illustrates. An independent verifier re-derives the label of a detection from the pixels inside its predicted box, and the disagreement between the verifier and the detector becomes the acquisition signal. Two properties follow immediately. The fusion problem disappears instead of being solved, because the crop taken from a predicted box already entangles the box and the label. A box that is displaced, that covers mostly background, or that is fired on nothing at all produces a crop that does not depict the claimed category, exactly as a correct box with a wrong label does. Verifying the crop therefore returns a single scalar in which label errors, localization errors and background false positives are already combined, and no relative weighting is introduced because none is needed. Confident errors also become visible, since the verifier never sees the detector confidence and is trained only to map pixels to labels, so its disagreement is not attenuated by the certainty of the detector.

Disagreement between models is of course the classical query by committee signal \cite{seung92,freund97}, whose deep instantiations appear among our baselines. The arrangement here differs in that the second model is not a second detector but a verifier solving an easier problem on an input the detector has already produced, which is what keeps the box out of the score as a separate term. Section \ref{sec2.5} sets out the comparison in full. Table \ref{tab:paradigm} places the proposal against the families it departs from.

\begin{table}[t]
\caption{Every established family reads its signal out of the detector being improved, which is what forces a label term to be combined with a box term. Generative verification is the only entry that reads the signal out of an independent model, and the only one for which the two failure modes need no weighting.}
\centering
\setlength{\tabcolsep}{5pt}
\renewcommand{\arraystretch}{1.2}
\resizebox{\textwidth}{!}{%
\begin{tabular}{llll}
\toprule
Family & Representative & Signal source & Box and label \\
\midrule
Output uncertainty & Entropy \cite{ref03}, GMM \cite{ref9}, EDL \cite{ref12} & detector output head & separate, hand-weighted \\
Learned proxy & LLAL \cite{ref5} & auxiliary head on features & implicit in predicted loss \\
Feature geometry & Core-set \cite{ref1}, CDAL \cite{ref8} & detector features, context & not modeled \\
Perturbation & Feature-mixture \cite{zhang} & detector output under perturbation & implicit \\
Decoupled query & DeLR \cite{delr} & detector, two queries & deliberately separated \\
Committee disagreement & QBC \cite{seung92}, Ensemble \cite{ref6}, MC-dropout \cite{ref18} & several detectors or passes, each at the full task & separate terms \\
Refined acquisition & PPAL \cite{ppal24}, MGRAL \cite{liang2026performance} & detector, reweighted or reinforced & separate terms \\
\midrule
\textbf{Generative verification} & \textbf{ours} & \textbf{independent generative model on the crop} & \textbf{jointly, by construction} \\
\bottomrule
\end{tabular}}
\label{tab:paradigm}
\end{table}

A verifier could in principle be an ordinary classifier, and the question of what a generative model adds deserves a direct answer. Part of the answer is not about normalization. Many crops produced by a detector contain no object at all, so our verifier carries an explicit background class, and a discriminative head with the same background class could likewise report that a crop depicts no foreground object. The advantage we claim lies elsewhere. A classifier returns a point estimate. One forward pass yields one score, and that score does not reveal how stable the judgement behind it is, because a single number conflates the strength of the evidence with the calibration of the head reporting it. Two crops receiving the same score may differ entirely in how firmly the evidence determines their label, and an acquisition function reading only the score cannot separate them.

A diffusion verifier is stochastic by construction. Its reverse process starts from a fresh Gaussian sample, so it can be run repeatedly on one crop, and the repeated generations trace out a distribution over label representations rather than a single point. The concentration of that distribution reports how firmly the evidence determines the label, which is a strictly richer object than any single score. Two properties make the variation useful here. It originates in the sampling process of the verifier, so it stays independent of the detector confidence and preserves the independence the arrangement relies on. And it lives in label representation space, where the distance from a generation to the claimed target is directly interpretable as agreement, with no intermediate calibration step. The randomness that makes a diffusion model unsuitable as a classifier is, on the present reading, exactly the mechanism that makes it a good verifier, and the reversal is the conceptual core of our design.

Concretely we build a conditional diffusion model whose diffusion target is a label representation rather than an image and which is conditioned on the crop. Prior diffusion models for discriminative tasks \cite{ref202,ref228} also diffuse a response variable conditioned on covariates, but both collapse the generations back into a softmax and then classify, discarding the spread that we consume. We do not classify. We measure the distance between each generation and the claimed target, run the reverse process several times per detection, and rank unlabeled images by a function of the resulting agreement count and the detector confidence.

We claim the following contributions.

\begin{itemize}[leftmargin=*]
\item \textbf{A different source for the acquisition signal.} We introduce generative verification, and Table \ref{tab:paradigm} positions it against the families that read their signal out of the detector.
\item \textbf{One scalar for both failure modes.} Verifying the crop makes the weighting of a label term against a box term unnecessary, and exposes the confidently wrong detections that a self-derived signal cannot see.
\item \textbf{A verifier that returns a distribution rather than a point estimate.} We diffuse label representations conditioned on the crop and exploit the stochasticity of the reverse process by generating several times per detection, so that the concentration of the generations, and not a single score, drives the ranking. Prior diffusion models for discriminative tasks \cite{ref202,ref228} collapse their generations back into a softmax and discard exactly that spread.
\item \textbf{Empirical validation.} On PASCAL VOC and MS-COCO the signal outperforms output-uncertainty, feature-geometry, perturbation and ensemble baselines, with the largest margins in the early rounds the arrangement is built for.
\end{itemize}
\section{Related Work}
\label{sec2}

\subsection{Diffusion models}
\label{sec2.1}
Comprehensive reviews of diffusion models are available in \cite{ref229,ref230,ref231}, and we restrict attention to the work our design builds on. Ho et al. \cite{ref216} introduced the denoising diffusion probabilistic model (DDPM), which destroys data to noise along a forward Markov chain and recovers it along a reverse chain, and our forward and reverse algorithm follows theirs. Song et al. \cite{ref217} cast the two directions as a stochastic differential equation and its reverse, and \cite{ref218} perturbed data at several noise levels, estimated the scores jointly and sampled with Langevin dynamics. Rombach et al. \cite{ref201} moved diffusion into the latent space of an autoencoder to reduce cost and introduced cross-attention layers that turn a diffusion model into a generator for general conditioning inputs, and we borrow their conditioning mechanism. Diffusion models have since been applied across image generation \cite{ref216,ref217,ref218,ref309,ref312}, super-resolution \cite{ref210,ref315}, inpainting \cite{ref220,ref310,ref311}, editing \cite{ref207}, image-to-image translation \cite{ref222}, and discriminative downstream tasks that read their latent representations, including segmentation \cite{ref223}, classification \cite{ref224} and anomaly detection \cite{ref225}, as well as beyond vision \cite{ref203,ref204,ref205,ref206,ref208,ref209,ref226}.

Two works are close enough to require explicit separation. Han et al. \cite{ref202} proposed classification and regression diffusion models, which pair a denoising conditional generative model with a pre-trained conditional mean estimator to predict the distribution of a response variable given its covariates, and which for classification diffuse a one-hot vector treated as a class prototype. Their instance-level confidence assessment exploits the stochasticity of the generative output, as ours does. Du et al. \cite{ref228} adopted the diffusion classifier of \cite{ref202} for active domain adaptation, generating $N$ predictions per unlabeled sample, converting them to probabilities with a softmax, selecting the two most predicted categories and applying a t-test to the two groups of scores. Both works therefore reduce their generations to class probabilities before making a decision, so what reaches the decision is a summary of the generations rather than their spread. We keep the spread, measuring each generation as a distance from the claimed target and letting the concentration across generations drive the ranking, as set out in Section \ref{sec1}.

\subsection{Signals read from the detector}
\label{sec2.2}
The literature is large, and we group it by the signal that drives the acquisition rather than by date, since the grouping is what locates our own contribution.

The largest group reads an informativeness score out of the detector outputs. Roy et al. \cite{ref03} exploited the disagreement between convolutional layers of the detector. Aghdam et al. \cite{ref3} aggregated pixel-level scores into an image-level score for pedestrian detection. Kao et al. \cite{ref4} combined localization and classification uncertainty. Yoo et al. \cite{ref5} predicted the loss of each unlabeled sample and ranked by the prediction. Li et al. \cite{ref09} aggregated information across multiple outputs and boxes while addressing class imbalance. Yu et al. \cite{ref11} required consistency of both the box and the predicted class distribution. Liu et al. \cite{ref010} estimated the expected gradient of an unlabeled sample to compute its influence. The common limitation is a dependence on the calibration of the very detector being improved, so an error made confidently by the detector is invisible to the score derived from it. A second group replaces the point estimate of the head with a distribution. Choi et al. \cite{ref9} substituted Gaussian mixture models for the output layers and read uncertainty from their parameters, giving the GMM and Prob baselines we compare against. Park et al. \cite{ref12} applied evidential deep learning with hierarchical uncertainty aggregation. Tang et al. \cite{ref31} combined dynamic architecture adaptation with Dirichlet calibration and clustering-based sampling. Schmidt et al. \cite{ref21} surveyed ensemble-based uncertainty estimators against their computational cost. We share the goal of a better calibrated signal but obtain the distribution from repeated stochastic generations rather than from a parametric output model.

\subsection{Diversity, instance-level and weakly supervised criteria}
\label{sec2.3}
A third group selects for coverage of the data distribution instead of for difficulty. Agarwal et al. \cite{ref8} used an information-theoretic distance to capture spatial and semantic diversity. Kothawade et al. \cite{ref020} applied submodular mutual information to target rare slices. Wu et al. \cite{ref24} combined instance-level uncertainty with diversity through entropy-based non-maximum suppression and a diverse prototype policy. The signal is complementary to ours rather than competing, and our results on VOC07+12 show methods of the kind overtaking us in the later rounds. Yuan et al. \cite{ref10} formulated the problem as multiple instance learning with instance uncertainty re-weighting and discrepancy learning. Brust et al. \cite{ref2} took an incremental learning route, and Wang et al. \cite{ref25} mined active samples in a way that also handles unseen classes. Desai et al. \cite{ref05} queried weak annotations adaptively, and in \cite{ref06} accounted for the number of ground truth boxes as a proxy for annotation effort. Zhang et al. \cite{zhang} probed the robustness of a detection by mixing its features with base representations at a small ratio and treated non-robust instances as informative, which is the closest baseline in spirit to ours since it also perturbs a detection and observes the stability of the outcome. Our work differs in that the perturbation is the stochasticity of a generative reverse process and the observation is made in label representation space rather than in feature space. We note that \cite{zhang} is our own prior work.

\subsection{Recent developments and adjacent paradigms}
\label{sec2.4}
Several works pair active learning with semi-supervised \cite{ref22,ref27,ref32} or weakly supervised learning \cite{ref23,ref30}. Su et al. \cite{ref7} unified adversarial domain alignment with importance sampling for the cross-domain setting, and Tang et al. \cite{ref28} transferred box-level supervision from a resource-rich source domain. Haussmann et al. \cite{ref6} reported a production system for autonomous driving, which supplies our Ensemble baseline. Application-specific studies cover salient object detection \cite{ref22,ref26}, aerial imagery \cite{ref29} and unmanned aerial vehicle imagery \cite{ref33}. Feng et al. \cite{ref34} contributed a benchmark under a shared training and testing protocol. Wan et al. \cite{midl23} extended multiple instance active object detection with differentiation learning, giving the journal version of \cite{ref10}. Yang et al. \cite{ppal24} proposed a plug and play scheme decoupling difficulty calibrated uncertainty from category conditioned diversity so that the acquisition function transfers across detectors without retuning. Zhang et al. \cite{delr} decoupled the two queries, as discussed in Section \ref{sec1}. Hekimoglu et al. \cite{noris} addressed the redundancy of instance level scores within an image. Sharma et al. \cite{sharma2026portable} pursued portability across detector families, and Liang et al. \cite{liang2026performance} used reinforcement learning driven by downstream detector performance rather than by a hand-designed uncertainty. Wang and Zhao \cite{umd25} combined uncertainty with diversity for indoor 3D detection, and Lin et al. \cite{saood26} addressed sparsely annotated oriented detection in remote sensing. Kassem Sbeyti et al. \cite{oss25} argued that reported gains in the area are sensitive to protocol details and released tooling to standardize the comparison. Menke et al. \cite{menke24} and Sokolov et al. \cite{sokolov25} studied the problem under domain shift and at the level of region of interest embeddings. Garcia et al. \cite{garcia23} surveyed a decade of the intersection between active learning and object detection.

We situate our own contribution against the recent work explicitly. Methods such as \cite{ppal24,delr,liang2026performance} improve the acquisition function while keeping the detector as the sole source of evidence, whereas ours introduces an external generative verifier. The two directions are compatible rather than competing. Our experimental comparison moreover predates \cite{sharma2026portable}, so it establishes our standing against the criteria listed in Section \ref{sec4.1} and not against the current frontier. All of these refine the acquisition function while keeping the detector as the source of evidence, and are therefore complementary to generative verification rather than alternatives to it. Concurrently with the present work, Liang et al. \cite{liang2026performance} and Sharma et al. \cite{sharma2026portable} report criteria on the same SSD and VOC07+12 protocol, the former driving selection by measured downstream detector gain and the latter by portability across detector families. Both appeared as preprints while the present work was under review and both retain the detector as the source of evidence.

The acquisition strategies used for detection inherit their vocabulary from active learning at large, surveyed by Ren et al. \cite{ren2022survey}, and two threads bear directly on the detection setting. The first is class imbalance under a fixed budget. Zhang et al. \cite{zhang2018online} studied active learning under severe class imbalance and a capped budget. A detector faces that situation in acute form, since background candidates outnumber foreground ones by orders of magnitude, which is why our verifier carries an explicit background class. The second is that the cost of an error is not uniform. Zhao et al. \cite{zhao2018adaptive} developed adaptive cost-sensitive online classification on the premise that different misclassifications carry different penalties, a premise our ranking function shares, since Equation \ref{eqn:un} deliberately promotes confidently wrong detections above uncertain ones. Active learning has also been carried into adjacent vision tasks where a prediction carries both a label and a geometry, such as visual tracking \cite{yuan2023active}, where the same difficulty of scoring two coupled predictions arises and is handled by combining separate terms. A complementary response to the annotation bottleneck is to exploit unlabeled images rather than to choose which ones to label. Ma et al. \cite{liu2023ambiguity} addressed the ambiguity of pseudo-labels in dense detection, and consistency-based schemes have been developed for oriented detection \cite{fu2024consistency}. The two families are orthogonal in principle, and combining a verifier with pseudo-labeling is attractive because the verifier already produces exactly the accept or reject decision that pseudo-label filtering needs.

\subsection{Relation to query by committee}
\label{sec2.5}
Using the disagreement between models as an acquisition signal is not new, and query by committee \cite{seung92,freund97} is its classical form. A committee of hypotheses consistent with the labeled data votes on each unlabeled sample, and the samples that split the vote are queried. Its modern deep instantiations are exactly the ensemble \cite{ref6} and Monte Carlo dropout \cite{ref18} criteria that our comparison already includes, and Roy et al. \cite{ref03} apply the same idea within a single network by contrasting the predictions of different convolutional layers. Since the family is represented in Tables \ref{tab:voc07} and \ref{tab:mscoco}, we are able to compare against it empirically rather than only in argument.

Three differences separate generative verification from a committee. The first is specific to detection. A committee here is composed of detectors, so every member predicts a box as well as a label and the disagreement must be defined over both, which reinstates the fusion problem rather than removing it. Our verifier is not a second detector but solves a strictly easier problem, recovering a label from a crop the detector has already produced, so the box enters through the crop instead of through a second term. The second difference is one of symmetry, since a committee has no privileged member and reads the variance among peers, whereas here the detector proposes and the verifier checks without ever seeing the detector confidence. The third is the origin of the variation, which a committee obtains from several models and we obtain from the stochastic reverse process of one. To our knowledge no prior work in any of the groups above uses a generative model to verify detections for acquisition, which is the gap our method addresses.

\section{Methodology}
\label{sec3}

\begin{figure*}[t]
\centering
\includegraphics[width=\linewidth]{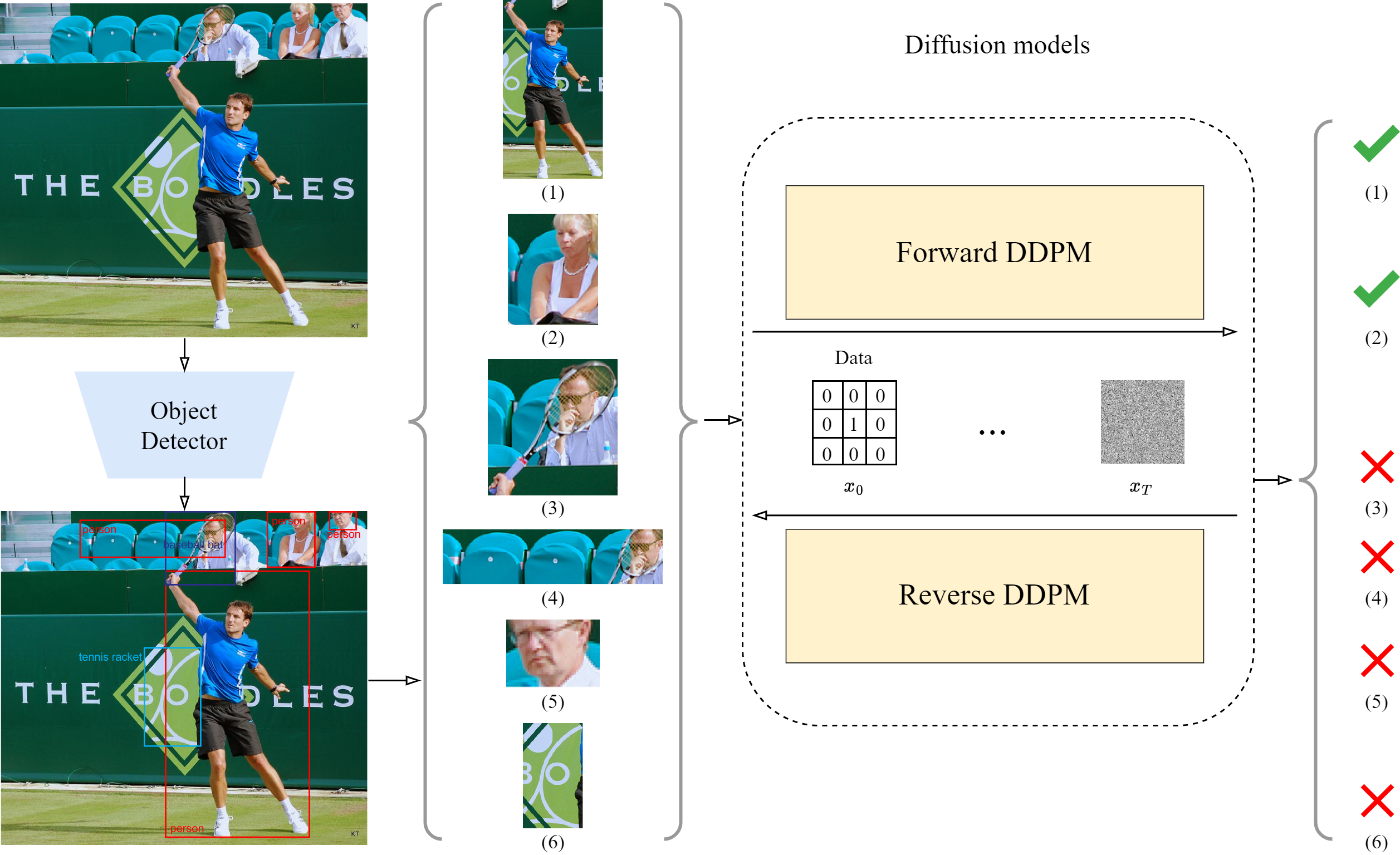}
\caption{The pipeline, and the seven outcomes it has to cover. The detector produces detections, each is cropped, and the crop alone is passed to the verifier. A single verification therefore covers what a conventional criterion has to reach through a separate label term and box term. In the illustration the verifier confirms the first two detections, which are accurate or carry only a slight offset, and rejects the third to the sixth, which carry a wrong label, cover mostly background, cover a fragment, or fire on nothing at all.}
\label{fig:framework}
\end{figure*}

\subsection{Reformulation}
\label{sec3.1}
Active learning for object detection has to score a prediction that carries two parts, a label and a box, whereas active learning for classification scores a label alone. The prevailing response, as set out in Section \ref{sec1}, is to estimate an uncertainty for each part and combine them. We take a different route, which rests on a single observation. The crop taken from a predicted box, read together with the predicted label, already carries the information about both. If the predicted label is wrong or the predicted box is inaccurate, the crop does not depict the claimed object. Verifying the crop is therefore sufficient, and no separate localization term is required.

The left half of Figure \ref{fig:framework} enumerates the outcomes an arbitrary detector produces, and it is worth checking the reformulation against each. A detection may be accurate in both label and box, as in the first sample. It may carry the correct label with a small positional offset, as in the second, either covering some region that does not belong to the object or missing some region that does, which does not materially affect performance and should be confirmed. It may localize correctly but assign the wrong label, as in the third, where a tennis racket is predicted as a baseball bat. It may cover part of an object together with a large area of background, or only a small portion of an object, or fire on background alone. Finally, the detector may miss an object entirely.

Under our formulation the first six outcomes reduce to the single question of whether the crop is consistent with the claimed label, and the intended behavior of the verifier follows. Accurate detections and detections with a slight offset are confirmed, since the crop still depicts the claimed object. Detections carrying the wrong label are rejected, since the crop depicts a different category. Detections whose box covers mostly background, or only a fragment of an object, and detections fired on pure background fall to the background class, since the crop carries too little evidence for any foreground category. A single verification thus covers the six outcomes that a conventional criterion has to reach through two separate uncertainty terms.

\subsection{Pipeline}
\label{sec3.2}
Figure \ref{fig:framework} shows the pipeline. A detector is trained on the labeled set and run on the unlabeled pool, each retained detection is cropped and passed to the verifier, and the disagreement between the generated label representation and the detector label becomes a detection-level score. Scores are summed into an image-level score, the highest scoring images are annotated, and the detector is retrained. Sections \ref{sec3.3} and \ref{sec3.4} specify the verifier and the condition it consumes, Section \ref{sec3.5} the acquisition procedure, and Algorithm \ref{alg} states it in full.

We instantiate the detector with SSD \cite{ref13}, under which the methods we compare against report their results \cite{ref03,ref4,ref5,ref8,ref9,ref10,ref12}, so that the comparison isolates the acquisition signal from the strength of the detector. We make no change to the model and train it exactly as in \cite{ref13}, with the VGG16 backbone \cite{vgg} used by previous work \cite{ref9,ref12}. Nothing in the arrangement depends on the choice, since the verifier consumes only crops and labels.

\subsection{The verifier as a conditional diffusion model}
\label{sec3.3}
We adopt the forward and reverse algorithm of \cite{ref216}. The forward process is
\begin{equation}
\label{eqn:forwardprod}
q(x_{1:T}|x_0) := \prod_{t=1}^{T} q(x_t|x_{t-1}),
\end{equation}
with
\begin{equation}
\label{eqn:forward}
q(x_t|x_{t-1}) := \mathcal{N} (x_t; \sqrt{1-\beta_t}x_{t-1}, \beta_t\textbf{I}),
\end{equation}
Here $x_0$ is the clean data sample, in our case a label representation rather than an image. The corrupted sample after $t$ forward steps is $x_t$, the total number of steps is $T$, and $x_{1:T}$ denotes the sequence $x_1, \ldots, x_T$. The symbol $q$ denotes the fixed forward distribution, $\mathcal{N}(\cdot; \mu, \Sigma)$ a Gaussian with mean $\mu$ and covariance $\Sigma$, $\beta_t$ with $t\in [1, T]$ the noise coefficients of the Markov transition kernels, and $\textbf{I}$ the identity matrix. The forward process gradually corrupts the input to Gaussian noise.

\begin{figure}[t]
\centering
\includegraphics[width=0.85\linewidth]{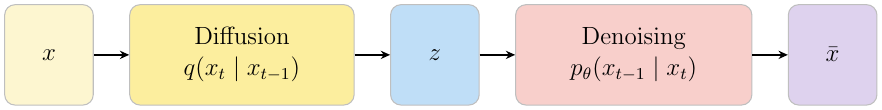}
\caption{The forward and reverse processes we adopt \cite{ref216}. Our verifier departs from the standard formulation in one respect only, namely what is diffused, which for us is a label representation rather than an image.}
\label{fig:ddpm}
\end{figure}

The reverse process is
\begin{equation}
\label{eqn:reverse}
p_\theta(x_{t-1}|x_t) := \mathcal{N} (x_{t-1}; \mu_\theta(x_t, t), \sigma^2_t\textbf{I}),
\end{equation}
where $p_\theta$ denotes the learned reverse distribution, $\theta$ the parameters of the denoising network, $\mu_\theta(x_t, t)$ the learnable mean predicted at step $t$, and $\sigma^2_t$ the variance at step $t$, which we keep fixed rather than learned. Over the whole trajectory,
\begin{equation}
\label{eqn:reverseprod}
p_\theta(x_{0:T}) := p(x_T)\prod_{t=1}^{T} p_\theta(x_{t-1}|x_t).
\end{equation}
Training fits $\theta$ so that $p_\theta(x_{0:T})$ approximates the forward trajectory. The objective reduces to the simplified form
\begin{equation}
\label{eqn:objective}
L_{diff}(\theta) := \mathbb{E}_{t, x_0, \epsilon}\left[||\epsilon - \epsilon_\theta(\sqrt{\overline{a}_t}x_0 + \sqrt{1 - \overline{a}_t}\epsilon, t)||^2\right],
\end{equation}
where $t$ is drawn uniformly from $1$ to $T$, $\epsilon \sim \mathcal{N}(0, \textbf{I})$ is the noise actually added in the forward process, $\epsilon_\theta$ is the denoising network that predicts the noise from the corrupted sample and the step index, $\mathbb{E}$ denotes expectation over the indicated variables, $a_t := 1 - \beta_t$ and $\overline{a}_t := \prod_{s=1}^t a_s$. Figure \ref{fig:ddpm} illustrates the model.

\textbf{Conditioning.} Following \cite{ref201} we build the denoising network on the UNet architecture \cite{ref301}. The condition enters through cross-attention at selected middle layers,
\begin{equation}
\label{eqn:attention}
{\rm Atte}(ir, cr) := {\rm softmax}\left(\frac{(W_Q\cdot ir)(W_K\cdot cr)^T}{\sqrt{d}}\right)\cdot (W_V\cdot cr),
\end{equation}
where $ir$ is a flattened intermediate representation of a middle layer of the UNet, $cr$ is the flattened conditional representation described in Section \ref{sec3.4}, $W_Q$, $W_K$ and $W_V$ are learnable projection matrices, and $d$ is the dimension of the projected query and key. We stress that the softmax in Equation \ref{eqn:attention} normalizes attention weights inside the network and is unrelated to a classification softmax over categories, which our pipeline never applies.

\begin{figure*}[t]
\centering
\includegraphics[width=\linewidth]{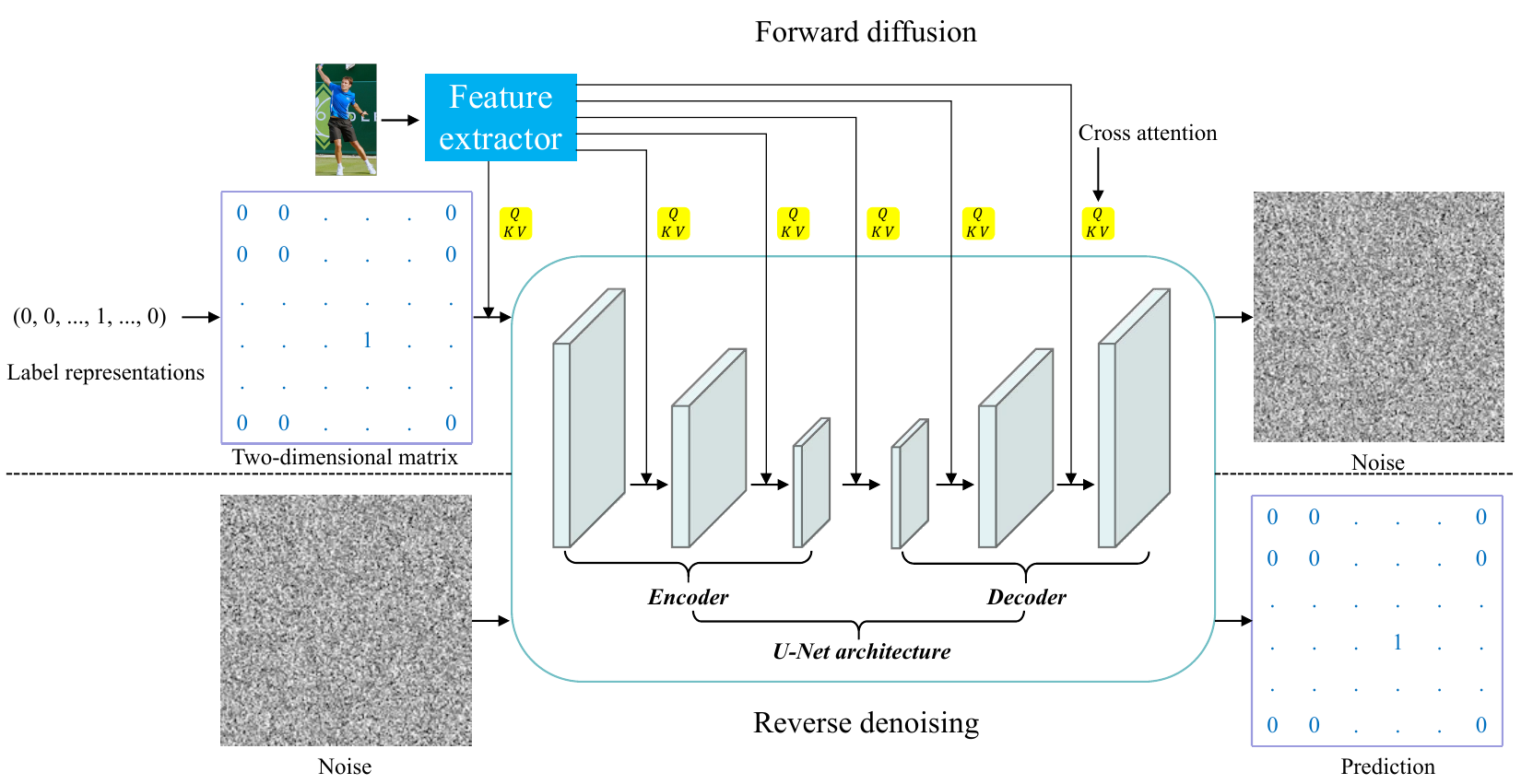}
\caption{Architecture of the verifier. Two choices distinguish it from a diffusion model for image synthesis. The diffusion target is a label representation, obtained by padding a one-hot class vector and reshaping it to a square grid, and the condition is a feature of the cropped detection injected through cross-attention. Repeated runs of the reverse process on one crop yield the distribution whose concentration our ranking function consumes.}
\label{fig:architecture}
\end{figure*}

\textbf{Label representations as the diffusion target.} Figure \ref{fig:architecture} shows the resulting model. The input of a diffusion model is conventionally an image. Ours generates label representations, so the diffusion target is a label representation and every category owns one. When the detector claims a crop is a person, the label representation of person is the target. We use one-hot vectors. Any representation is admissible in principle, since all that is required is a distance separating agreement from disagreement, but a more entangled representation makes the generation harder to learn, so we adopt the simplest option.

To reuse a standard image-space UNet without modification we pad the one-hot vector to $2^{2n}$ dimensions and reshape it to $2^n \times 2^n$, using 64 dimensions for the 20 foreground classes of Pascal VOC and 256 for the 80 of MS-COCO. The design inherits a well tested architecture unchanged, and a one-dimensional denoiser over the raw class vector is an equally valid instantiation.

\subsection{Condition design}
\label{sec3.4}
The verifier is conditioned on the cropped detection. Rather than feeding raw pixels we encode the crop with a model pre-trained on ImageNet \cite{imagenet} and use the resulting vector as the condition. The encoder is CAFormer \cite{ref302}, specifically the B36 version at $224 \times 224$ resolution, pre-trained on ImageNet-21K and fine-tuned on ImageNet-1K. We take the final layer output, a 1000-dimensional vector, as the condition.

A verifier under our arrangement therefore comprises two parts, an encoder that maps pixels to a representation and a generative head that maps the representation to a label representation. The division is intrinsic to the arrangement, since any verifier must see the pixels somehow. The role reserved for the generative head is not to supply the semantics but to turn a single deterministic encoding into a distribution over label representations, because it is the spread of the repeated generations that the ranking function consumes.

\subsection{Acquisition algorithm}
\label{sec3.5}
The workflow begins by drawing a small subset of the unlabeled pool at random and having it annotated, which suffices to train a preliminary detector. From there each cycle proceeds as follows.

An image yields many detections, and we keep at most 3 per image, namely those of highest confidence among the detections that clear the selection threshold. An image with fewer than 3 such detections contributes all of them. Pascal VOC carries under 3 annotated objects per image on average, so the budget covers a typical VOC image almost entirely. MS-COCO averages around 7, so there the budget inspects only the most confident part of each image. The asymmetry is a compute driven approximation.

For each retained detection we crop the predicted box and encode the crop with CAFormer. The reverse process is then run on random Gaussian noise, conditioned on that encoding, and returns a generated label representation, which we flatten. A generation is counted as agreeing with the detector when
\begin{equation}
\label{eqn:agree}
|g(i) - 1| \leq \delta,
\end{equation}
where $g$ is the flattened generation, $i$ is the label predicted by the detector, $g(i)$ is the $i$-th element of $g$, and $\delta = 0.15$ throughout. The criterion asks that the generated value at the claimed category be within $\delta$ of the target value 1. Otherwise the generation disagrees.

Because the reverse process is stochastic we repeat it $N$ times per detection and let $n$ be the number of agreeing generations. The detection-level score is
\begin{equation}
\label{eqn:un}
uncert = \begin{cases}
10^{6} \cdot score + 10^{8}, & \text{if} \ n=0 \\
-10^{3} \cdot score,  & \text{if} \ n=N \\
\frac{N-n}{N}, & \text{otherwise,}
\end{cases}
\end{equation}
where $score$ is the detector confidence of the detection. When the fully disagreeing case $n=0$ is too strict to yield enough candidates, the first branch can be relaxed to $n\leq 1$.

Equation \ref{eqn:un} defines a bucketed order, and reads most easily as three groups separated by construction. Detections on which the verifier fully disagrees with the detector form the top group and are ordered within it by $score$, so that a more confident error ranks higher, since the detector being confidently wrong is precisely the case the annotation budget should address. Detections on which the verifier fully agrees form the bottom group and receive a negative value ordered by $-score$, so that confidently correct detections are pushed furthest down. The remaining detections, on which the verifier is itself divided, are ordered by the fraction of disagreeing rounds, so that the score decreases monotonically as the agreement count $n$ grows. The constants keep the three groups apart. With the selection thresholds of Section \ref{sec4.1} the confidence $score$ lies in $[0.1, 1]$, so the top group falls above $10^8$, the middle group within $(0, 1)$ and the bottom group within $[-1000, -100]$. The three ranges are disjoint, so no detection can overtake one from a higher group whatever its confidence. An implementation may equivalently sort by group index first and by the corresponding within-group key second. Because the separation between groups is wider than the spread within any of them, summing the scores of an image ranks it first by how many of its detections fall in the top group and only then by how confident those errors are, so an image carrying two confident errors outranks one carrying a single error however confident.

The image-level score is the sum over the retained detections of the image. Algorithm \ref{alg} summarizes the procedure.

\begin{algorithm}[t]
\setstretch{1.0}\small
\caption{Active learning of object detection by generative verification. The detector is retrained from scratch in every cycle, and the verifier is refitted on the enlarged labeled set, so neither model carries state across cycles.}
\label{alg}
\KwIn{Detector $Det$, verifier $DM$, encoder $Enc$, labeled set $L$, unlabeled pool $U$, budget $k$, generations $N$, cycles $R$, thresholds $\tau$ and $[\tau_{lo}, \tau_{hi}]$.}
\KwOut{Detector trained on the final labeled set.}
\For{$r \gets 1$ \KwTo $R$}{
Train $Det$ on $L$ from scratch.\\
$B \gets$ detections of $Det$ on $L$ with confidence in $[\tau_{lo}, \tau_{hi}]$.\\
$F \gets$ ground truth crops of $L$, rare classes repeated.\\
Train $DM$ on $F \cup B$.\\
$P \gets \{\}$.\\
\For{$u \in U$}{
$D \gets$ up to 3 highest-confidence detections of $Det$ on $u$ above $\tau$; \ \ $s \gets 0$.\\
\For{$d \in D$}{
$cr \gets Enc(\text{crop of } d)$.\\
Draw $N$ generations from $DM$ given $cr$.\\
$n \gets \#$ generations meeting Eq. \ref{eqn:agree}.\\
$s \gets s + uncert(n, score_d)$ by Eq. \ref{eqn:un}.\\
}
$P \gets P \cup \{(u, s)\}$.
}
Move the top-$k$ of $P$ from $U$ to $L$ after annotation.
}
Train $Det$ on the final $L$.
\end{algorithm}

\section{Experiments}
\label{sec4}
\subsection{Empirical settings}
\label{sec4.1}
\textbf{Datasets.} \textit{Pascal visual object classes (VOC)} \cite{pascal} and \textit{Microsoft common objects in context (MS-COCO)} \cite{coco} carry instances in 20 and 80 foreground classes respectively, and counting the background class the label representations produced by our verifier span 21 and 81 categories. The two datasets contain around 16000 and 82000 labeled images. For Pascal VOC we use two settings, training and testing on VOC 2007, and training on VOC 2007 plus the VOC 2012 training set while testing on the VOC 2007 test set, abbreviated VOC07 and VOC07+12. For MS-COCO we train on train2014 and test on val2017, following \cite{ref9}.

\textbf{Evaluation.} We report mean Average Precision at 0.5 IoU (mAP50), which previous work adopts \cite{ref9,ref12}. The computation differs between the two benchmarks, with the VOC07 11-point method for Pascal VOC and the MS-COCO evaluation method for MS-COCO, and we follow each.

\textbf{Comparison methods.} Eleven criteria derive their score from a single detector. \textit{Random} selects images at random. \textit{Entropy} \cite{ref03} takes the entropy of the detection boxes, \textit{Core-set} \cite{ref1} a cover of the remaining pool, and \textit{LLAL} \cite{ref5} a learned prediction of the target loss. \textit{CDAL} \cite{ref8} contributes two entries, a core-set variant (CS) and a reinforcement learning variant (RL). \textit{MI-AOD} \cite{ref10} uses multiple instance learning, \textit{EBAL} \cite{ref24} entropy under a progressive diversity constraint, and \textit{EDL} \cite{ref12} evidential deep learning. \textit{PPAL} \cite{ppal24} decouples difficulty-calibrated uncertainty from category-conditioned diversity. \textit{Feature-mixture} \cite{zhang} probes the robustness of detections under feature mixture, and is our own prior work. Four further criteria use several detectors or several stochastic passes, namely \textit{MC-dropout} \cite{ref18} at 25 and 50 passes, \textit{Ensemble} \cite{ref6}, and \textit{GMM} and \textit{Prob} \cite{ref9}. \textit{EBAL} and \textit{PPAL} are reported on VOC07+12 only, the setting for which values under the shared protocol are available.

Our method adds a frozen ImageNet pre-trained encoder and a conditional diffusion model, both of which operate on cropped detections rather than on whole images. Throughout, the detector, the labeling budget and the evaluation protocol are held fixed, so that the comparison isolates the acquisition signal from every other factor.

\textbf{Detector training.} SSD was trained with SGD for 500 epochs, with 50 warmup epochs increasing the learning rate uniformly from $10^{-5}$ to $10^{-3}$, then 200 epochs at $10^{-3}$, 100 epochs at $10^{-4}$ and the remainder at $10^{-5}$. Batch size was 8, the input resolution $300 \times 300$, and VGG-16 was initialized from ImageNet pre-trained weights \cite{imagenet}. The selection threshold $\tau$ was 0.1 for Pascal VOC and 0.4 for MS-COCO.

\textbf{Repetitions.} Each experiment was repeated with three trials, and for the acquisition rounds we report the mean and standard deviation over them. Our protocol, namely SSD with a VGG16 backbone, VOC07+12 trainval for training, VOC07 test for evaluation and a 1k initial set enlarged by 1k per round, is the one under which most of the compared criteria already report results. Wherever a criterion has been evaluated under it we quote the published values rather than rerun the method, so that the numbers compare acquisition criteria and not reimplementations of them. The values for \textit{PPAL} and \textit{EBAL} on VOC07+12 are read from the published comparisons of \cite{sharma2026portable,liang2026performance}, whose two independent reports of \textit{PPAL} agree to within 0.05 mAP50 from the third round onward. The first round precedes any acquisition decision, since the initial subset is drawn at random and no criterion has yet been applied, so we report its mean alone as the starting point from which the acquisition rounds are measured. The reported deviations therefore span the two ends of a single phenomenon. A standard deviation of 0.00 indicates a criterion whose selection is deterministic under the protocol, so that the three trials agree on the same set of images and the only remaining variation is that of the detector training. At the other end, \textit{Random} carries the largest deviations of any entry, 0.60 and 0.45 on MS-COCO against 0.05 to 0.15 for the criteria, because its acquisition is redrawn in every trial, so its spread compounds the variance of the sampling with that of the training. Deviations should accordingly be compared within a row rather than across rows, and the deviation of a criterion measures how stable the trained detector is rather than how stable the criterion is.

\textbf{Verifier training.} The diffusion model was trained with AdamW for 500 epochs under a linear learning rate schedule from $10^{-4}$ to $10^{-5}$, batch size 256, with $T = 20$ and $\beta_t$ increasing linearly from $\beta_1 = 10^{-4}$ to $\beta_T = 0.02$. Foreground crops come from the ground truth boxes of the labeled images. The verifier also needs a background class, because several outcomes enumerated in Section \ref{sec3.1} produce crops containing no object. We mine background crops from the detector rather than from random boxes, so that they follow the same spatial prior as the detections the verifier will judge. We collect detections whose confidence falls in a narrow band just above the default output threshold of \cite{ref13}, namely $[0.01, 0.011]$. Boxes in the band are ones the detector emits without believing, and they are dominated by background and badly localized crops. We split the collected data 9:1 into training and validation, select the checkpoint by class-mean accuracy, and repeat samples from rare classes so that classes are roughly balanced. $N$ was 10. Label representations were 64-dimensional for Pascal VOC and 256-dimensional for MS-COCO, reshaped to $8 \times 8$ and $16 \times 16$.

\textbf{Verifier architecture.} The UNet has 1 input and 1 output channel, 192 base channels and 1 residual block per stage. Cross-attention is applied at the $4\times$, $2\times$ and $1\times$ downsampling feature maps. Three stages at $1\times$, $2\times$ and $4\times$ the base channel count are used for both encoder and decoder in reverse order. Head channels are 32, transformer depth is 1, and the conditional feature dimension is 1000.

\subsection{Results}
\label{sec4.2}

\begin{table}[t]
\caption{VOC07. Our criterion ranks first in both acquisition rounds against both the single and the multiple model-based groups. The first round precedes any acquisition decision, so its entries are the mean over three trials of a randomly drawn subset and are given for reference. The acquisition rounds report mean and standard deviation over three trials.}
\centering
\setlength{\tabcolsep}{8pt}
\begin{tabular}{lccc}
\toprule
\multirow{2}{*}{Method} & \multicolumn{3}{c}{mAP50 (\%)}\\
& 1st (2k) & 2nd (3k) & 3rd (4k)\\
\midrule
Random & 62.78 & 67.06$\pm$0.17 & 69.38$\pm$0.38 \\
Entropy \cite{ref03} & 62.43 & 66.85$\pm$0.12 & 68.70$\pm$0.18 \\
Core-set \cite{ref1} & 62.43 & 66.57$\pm$0.20 & 68.57$\pm$0.26 \\
LLAL \cite{ref5} & 62.47 & 67.02$\pm$0.11 & 68.90$\pm$0.15 \\
Feature-mixture \cite{zhang} & 62.37 & 67.76$\pm$0.03 & 69.98$\pm$0.01 \\
\midrule
MC-dropout \cite{ref18} & 62.43 & 67.10$\pm$0.07 & 69.39$\pm$0.09 \\
Ensemble \cite{ref6} & 62.43 & 67.11$\pm$0.26 & 69.26$\pm$0.14 \\
Prob \cite{ref9} & 62.91 & 67.61$\pm$0.17 & 69.66$\pm$0.17 \\
GMM \cite{ref9} & 62.43 & 67.32$\pm$0.12 & 69.43$\pm$0.11 \\
\midrule
\textbf{Ours} & 62.78 & \textbf{67.93$\pm$0.01} & \textbf{70.00$\pm$0.01} \\
\bottomrule
\end{tabular}
\label{tab:voc07}
\end{table}

\textbf{VOC07.} We drew two thousand images at random from the roughly 5000 of VOC07, then ran two acquisition rounds of one thousand images each, with random selection of three and four thousand images as a reference. Table \ref{tab:voc07} reports the outcome. The margin over Feature-mixture \cite{zhang} is 0.17 points at 3k and 0.02 at 4k. Both methods report a standard deviation of 0.01 at 4k, so the smaller gap is about two standard deviations of the run to run variation.

The lead narrows across rounds, as expected on a pool of such size. VOC07 offers only about 5000 training images, so by the third round a large fraction of it is already labeled and every strategy converges toward the same set. VOC07 is included because it is a standard point of comparison, and the two larger settings below separate the methods more sharply.

\begin{table}[t]
\caption{MS-COCO. The margin over the best baseline is about one mAP50 point in each acquisition round, an order of magnitude above the run to run deviation, and the largest of the three settings we study. The first round precedes any acquisition decision and is given for reference.}
\centering
\setlength{\tabcolsep}{8pt}
\begin{tabular}{lccc}
\toprule
\multirow{2}{*}{Method} & \multicolumn{3}{c}{mAP50 (\%)}\\
& 1st (5k) & 2nd (6k) & 3rd (7k)\\
\midrule
Random & 27.97 & 29.17$\pm$0.60 & 30.27$\pm$0.45\\
Entropy \cite{ref03} & 27.70 & 28.93$\pm$0.11 & 29.89$\pm$0.09\\
Core-set \cite{ref1} & 27.70 & 28.99$\pm$0.01 & 29.93$\pm$0.06\\
LLAL \cite{ref5} & 27.71 & 28.71$\pm$0.06 & 29.53$\pm$0.15\\
Feature-mixture \cite{zhang} & 27.10 & 28.60$\pm$0.00 & 29.60$\pm$0.00\\
\midrule
MC-dropout \cite{ref18} & 27.70 & 29.20$\pm$0.09 & 30.30$\pm$0.08\\
Ensemble \cite{ref6} & 27.70 & 29.03$\pm$0.07 & 30.02$\pm$0.06\\
Prob \cite{ref9} & 27.33 & 29.06$\pm$0.08 & 30.02$\pm$0.05\\
GMM \cite{ref9} & 27.70 & 29.28$\pm$0.05 & 30.51$\pm$0.12\\
\midrule
\textbf{Ours} & 27.97 & \textbf{30.33$\pm$0.05} & \textbf{31.53$\pm$0.05}\\
\bottomrule
\end{tabular}
\label{tab:mscoco}
\end{table}

\textbf{MS-COCO.} We drew five thousand images at random from the roughly 82000 of MS-COCO train2014 and ran two acquisition rounds of one thousand images each. Table \ref{tab:mscoco} reports the outcome, where our criterion leads both groups in both acquisition rounds and the strongest baseline is GMM \cite{ref9}.

The margins here exceed those on VOC07 for two reasons that the arrangement predicts. The pool is about sixteen times larger and is far from exhausted after two rounds, leaving room for an acquisition strategy to distinguish itself. MS-COCO also carries 80 foreground classes with heavy scale and occlusion variation, which produces exactly the mislabeled and badly localized detections the verifier is built to catch. The harder benchmark is therefore where the signal pays off most.

\begin{figure*}
\centering
\subfloat[]{\includegraphics[width=0.5\linewidth]{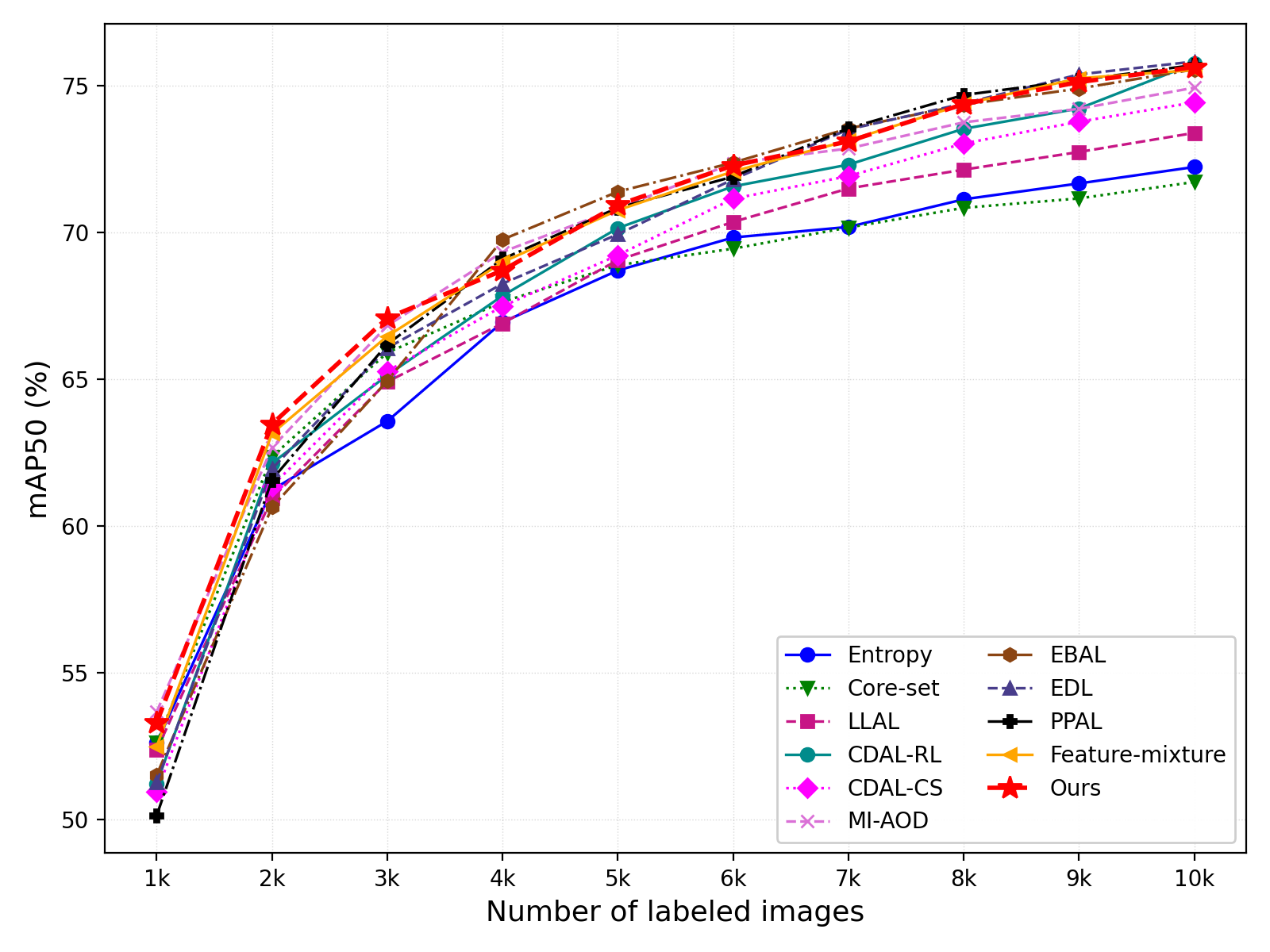}}
\hfill
\subfloat[]{\includegraphics[width=0.5\linewidth]{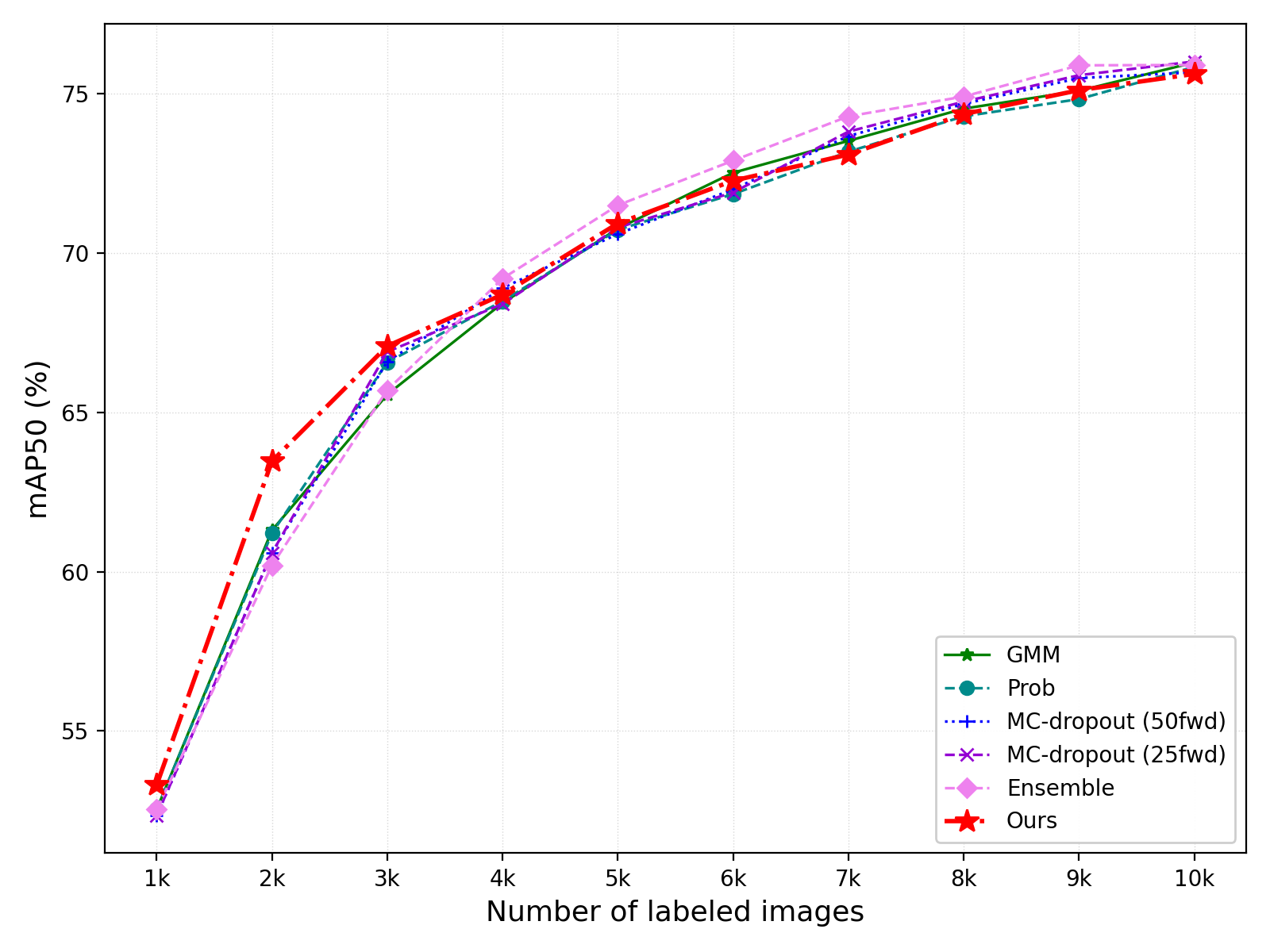}}
\caption{VOC07+12, where the advantage concentrates in the early rounds. Our criterion leads the most recent published criterion by 1.89 and 0.89 points in the first two acquisition rounds and is level with it from 8k onward, which is the behaviour the arrangement predicts, since confident detector errors grow scarce as the labeled set grows. (a) single model-based methods, (b) multiple model-based methods. All curves follow the protocol of Section \ref{sec4.1}. The values for PPAL and EBAL are read from the published comparisons of \cite{sharma2026portable, liang2026performance}, and those for the criteria of \cite{ref9} from the tabulated results in its supplementary material.}
\label{fig:voc0712}
\end{figure*}

\textbf{VOC07+12.} We drew one thousand samples at random from the roughly 16000 training images and ran nine acquisition rounds of one thousand images each, reaching ten thousand. Figure \ref{fig:voc0712}(a) compares against Entropy \cite{ref03}, Core-set \cite{ref1}, LLAL \cite{ref5}, CDAL-RL and CDAL-CS \cite{ref8}, MI-AOD \cite{ref10}, EBAL \cite{ref24}, EDL \cite{ref12}, PPAL \cite{ppal24} and Feature-mixture \cite{zhang}. On VOC07+12 the comparison spans 2018 to 2024, and it is the setting for which the most recent criteria have been reported under our own protocol.

Reading the late-round margins calls for a wider yardstick than the deviations in Tables \ref{tab:voc07} and \ref{tab:mscoco}, which measure repeated training under one protocol. Because several of the criteria plotted here appear in more than one published comparison, we can measure the spread of one and the same criterion across independent papers directly, and it averages about 0.4 mAP50. Against that yardstick the first two acquisition rounds separate our method from PPAL \cite{ppal24} while the last three do not. Over the remaining rounds our method stays within the band occupied by EBAL, EDL, Feature-mixture and CDAL-RL.

The early rounds are also the ones that matter most in practice, since the value of active learning lies in reaching a usable detector before the budget is spent. As the budget grows and confident detector errors grow scarce, methods that model diversity, such as CDAL-RL, or epistemic uncertainty, such as EDL, close the gap. The two signals are therefore complementary, and pairing a verifier with a diversity term is a direct extension.

Figure \ref{fig:voc0712}(b) compares against GMM \cite{ref9}, Prob \cite{ref9}, MC-dropout at 25 and 50 passes \cite{ref18} and Ensemble \cite{ref6}. Our method leads at 2k and 3k and stays within a narrow band of the group thereafter.

\subsection{Ablation}
\label{sec4.3}

\begin{table}[t]
\caption{Repeated generation is what the method depends on, VOC07. Ten generations per detection beat one in both acquisition rounds, by 0.20 and 0.33 points against deviations of 0.01 to 0.02. First-round entries are reference means, as in Table \ref{tab:voc07}.}
\centering
\setlength{\tabcolsep}{10pt}
\begin{tabular}{lccc}
\toprule
\multirow{2}{*}{Setting} & \multicolumn{3}{c}{mAP50 (\%)}\\
& 1st (2k) & 2nd (3k) & 3rd (4k)\\
\midrule
$N = 1$ & 62.78 & 67.73$\pm$0.02 & 69.67$\pm$0.02 \\
$N = 10$ & 62.78 & \textbf{67.93$\pm$0.01} & \textbf{70.00$\pm$0.01} \\
\bottomrule
\end{tabular}
\label{tab:voc07:ab}
\end{table}

\begin{table}[t]
\caption{Repeated generation is what the method depends on, MS-COCO. The margin is 0.26 and 0.36 points against deviations of 0.05, and the direction agrees with VOC07, so the effect holds across both datasets. First-round entries are reference means, as in Table \ref{tab:mscoco}.}
\centering
\setlength{\tabcolsep}{10pt}
\begin{tabular}{lccc}
\toprule
\multirow{2}{*}{Setting} & \multicolumn{3}{c}{mAP50 (\%)}\\
& 1st (5k) & 2nd (6k) & 3rd (7k)\\
\midrule
$N = 1$ & 27.97 & 30.07$\pm$0.05 & 31.17$\pm$0.05 \\
$N = 10$ & 27.97 & \textbf{30.33$\pm$0.05} & \textbf{31.53$\pm$0.05}\\
\bottomrule
\end{tabular}
\label{tab:mscoco:ab}
\end{table}

\textbf{Repeated generation is what the method depends on.} The argument of Section \ref{sec1} rests on the claim that the distribution traced by repeated generations carries more than any single generation does, so the decisive ablation sets $N = 1$ and leaves everything else unchanged. Tables \ref{tab:voc07:ab} and \ref{tab:mscoco:ab} report the comparison, which supports the premise the design rests on.

\textbf{Remaining settings.} The number of reverse steps is $T = 20$ and the agreement tolerance is $\delta = 0.15$. A padded one-hot vector is a far simpler diffusion target than a natural image, so it needs an order of magnitude fewer steps than image synthesis. The same pair of values serves a 20-class benchmark and an 80-class one without change, so the criterion transfers to a new dataset without a sweep, which is what an annotation campaign needs.

\section{Limitations and Future Work}
\label{sec5}
Four limitations bound what the present results establish.

\textbf{A single detector.} Every experiment uses SSD \cite{ref13} with a VGG16 backbone. Holding it fixed is what isolates the acquisition signal from the strength of the detector, but it leaves the signal untested on a detector that makes fewer and differently distributed errors. The verifier consumes only crops and claimed labels, so it attaches unchanged to a two-stage or a query-based detector, and carrying the protocol across detector families is the study we regard as most valuable. Because label representations are one-hot vectors, open-vocabulary detection is likewise out of reach until the target is replaced by a text embedding.

\textbf{The verifier is not decomposed.} Our verifier pairs a frozen ImageNet pre-trained encoder with a conditional diffusion head, and the experiments measure the pair rather than either part. The gap does not bear on the claim of Table \ref{tab:paradigm}, since an encoder is a component of any verifier rather than a confound external to one, but it does bear on the argument of Section \ref{sec1} that the head should be generative. Swapping the diffusion head for a lightweight discriminative head at a fixed encoder would settle it, and both outcomes are informative.

\textbf{Missed objects are outside the score.} A crop is produced only where the detector fires, so the score reads what a detector commits to and not what it passes over, although an image full of missed objects is worth annotating for the opposite reason. Pairing our score with a term sensitive to low detector recall would cover both, and the two compose without modification since they read disjoint evidence.

\textbf{The comparison stops at 2024.} The criteria we measure span the four families that Table \ref{tab:paradigm} contrasts with ours, which is what makes them the right set for a claim about where the signal should come from. Criteria published since \cite{delr, sharma2026portable, liang2026performance} refine the acquisition function while keeping the detector as the source of evidence, and semi-supervised detection \cite{liu2023ambiguity, fu2024consistency} exploits unlabeled images instead of choosing among them. Both operate on a different axis from a verifier, so combining them with one is the experiment their existence calls for.

\section{Conclusion}
\label{sec6}
We have argued that active learning for object detection should change where its acquisition signal comes from. Existing methods interrogate the detector under improvement. We instead verify each detection with an independent generative model, and take the disagreement between the two as the signal. The reformulation removes the weighting of a label term against a box term, and exposes the confident errors that a self-derived signal cannot see. We built the verifier as a conditional diffusion model over label representations, sampled repeatedly so that its spread rather than a single score drives the ranking. Experiments on PASCAL VOC and MS-COCO support the claim. Section \ref{sec5} sets out the limitations that bound the result.

\section*{CRediT authorship contribution statement}
\textbf{Licheng Zhang:} Conceptualization, Methodology, Software, Investigation, Visualization, Writing -- original draft. \textbf{Zheng Gong:} Supervision, Funding acquisition, Writing -- review and editing.

\section*{Declaration of competing interest}
The authors declare that they have no known competing financial interests or personal relationships that could have appeared to influence the work reported here.

\section*{Data availability}
All data used in this study are publicly available. PASCAL VOC \cite{pascal} and MS-COCO \cite{coco} are standard benchmarks distributed by their maintainers, and no new data were collected. The baseline values quoted in Tables \ref{tab:voc07} and \ref{tab:mscoco} and in Figure \ref{fig:voc0712} are taken from the published reports cited alongside them.

\section*{Acknowledgements}
Partial financial support for this study was provided by the National Natural Science Foundation of China (NSFC) through grants 42301468 and 42371457, the Fujian Province Natural Science Foundation, China (Grant No. 2023J01799), and the Xiamen Natural Science Foundation, China (Grant No. 3502Z20227048), along with contributions from the Jimei University Startup Fund (Grant ZQ2022031).

\section*{Declaration of generative AI and AI-assisted technologies in the writing process}
During the preparation of this work the authors used a large language model in order to improve the readability and language of the manuscript. The tool was not used to generate or alter any data, figures or experimental results. After using it the authors reviewed and edited the content as needed and take full responsibility for the content of the publication.

\bibliographystyle{elsarticle-num-names}
\bibliography{ref.bib}

~\\~\\~\\
\begin{wrapfigure}{l}{25mm}
\includegraphics[width=1in,height=1.25in,clip,keepaspectratio]{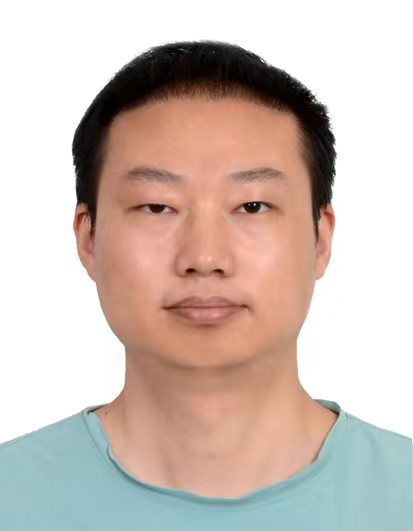}
\end{wrapfigure}\par
\textbf{Licheng Zhang} graduated from Yangzhou University in 2013 with a bachelor degree in Information and Computing Science. He graduated from Peking University in 2016 with a master degree in Intelligence Science and Technology. He is currently pursuing the Ph.D. degree with The University of Melbourne, Melbourne, VIC, Australia. His research interests lie in computer vision, deep learning and pattern recognition.

~\\~\\~\\
\begin{wrapfigure}{l}{25mm}
\includegraphics[width=1in,height=1.25in,clip,keepaspectratio]{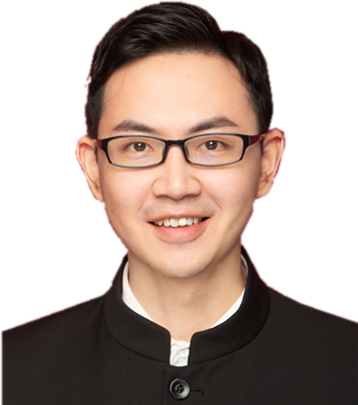}
\end{wrapfigure}\par
\textbf{Zheng Gong} received his Ph.D. in Communication and Information Systems from Xiamen University and was a visiting Ph.D. scholar at the University of Waterloo, Canada. He is currently a Research Fellow and Lecturer at the College of Computer Engineering, Jimei University, Xiamen. Previously, he worked as a Visual Algorithm Researcher at DJI Innovations in Shenzhen. His main research interests include computer vision and 3D perception for robotics, covering areas such as multi-sensor state estimation and optimization, point cloud semantic analysis, online calibration of multi-sensor fusion, and integrated perception-control networks based on reinforcement learning.

\end{document}